\documentclass{article}

\usepackage{arxiv}

\usepackage[utf8]{inputenc} 
\usepackage[T1]{fontenc}    
\usepackage{hyperref}       
\usepackage{url}            
\usepackage{booktabs}       
\usepackage{amsfonts}       
\usepackage{nicefrac}       
\usepackage{microtype}      
\usepackage{lipsum}

\usepackage{algorithm}
\usepackage[noend]{algpseudocode}

\usepackage{amsmath}

\usepackage{graphicx}

\title{Reactive Sample Size for Heuristic Search in Simulation-based Optimization}

\author{
  Manuel Dalcastagn\'e\\
  Department of Information Engineering and Computer Science\\
  University of Trento\\
  Povo, Italy 38123\\
  \texttt{m.dalcastagne@unitn.it}\\
   \And
  Andrea Mariello\\
  Department of Information Engineering and Computer Science\\
  University of Trento\\
  Povo, Italy 38123\\
  \texttt{andrea.mariello@alumni.unitn.it}\\
   \And
  Roberto Battiti\\
  Department of Information Engineering and Computer Science\\
  University of Trento\\
  Povo, Italy 38123\\
  \texttt{roberto.battiti@unitn.it}\\  
}

\begin{document}
\maketitle

\begin{abstract}
In simulation-based optimization, the optimal setting of the input parameters of the objective function can be determined by heuristic optimization techniques. 
However, when simulators model the stochasticity of real-world problems, their output is a random variable and multiple evaluations of the objective function are necessary to properly compare the expected performance of different parameter settings.
This paper presents a novel reactive sample size algorithm based on parametric tests and indifference-zone selection, which can be used for improving the efficiency and robustness of heuristic optimization methods. 
The algorithm reactively decides, in an online manner, the sample size to be used for each comparison during the optimization according to observed statistical evidence. 
Tests employ benchmark functions extended with artificial levels of noise and a simulation-based optimization tool for hotel revenue management. 
Experimental results show that the reactive method can improve the efficiency and robustness of simulation-based optimization techniques.
\end{abstract}


\section{Introduction}
\label{sec:intro}
It is well known that many real-world problems cannot be solved to optimality in acceptable (polynomial) CPU times \cite{ausiello2012}.
In fact, heuristic search algorithms are often used to obtain improving state-of-the-art results to complex problems in a reasonable amount of time.
The solutions found by these methods are not guaranteed to be optimal, but are suitable for many practical applications.

In the context of simulation-based optimization, the objective of heuristic algorithms is to find solutions to stochastic problems trough the intelligent use of algorithmic building blocks based on local search. 
The optimization starts from an initial solution of the simulation model, and then it iteratively improves the best solution through a search in its neighborhood.
\cite{amaran2016,battiti2017} review some of these approaches.
\cite{andradottir2006,olafsson2006} provide an analysis of the role that various heuristic optimization techniques have in simulation-based optimization.
Other surveys and books about simulation-based optimization are \cite{fu2005,hong2009,gosavi2014,hong2015,jian2015}. However, as highlighted by \cite{linz2017}, little attention has been given to the impact that the method used to estimate the objective function has on the performance of simulation-based optimization methods.
In fact, optimization techniques need to deal with the stochasticity of the objective function, by using sufficiently accurate estimators in comparisons done during the search.

As stated by \cite{boesel2000}, \cite{hong2009}, an established practice is to introduce a statistical analysis after the heuristic search by using ranking and selection (R\&S) algorithms.
This analysis aims at selecting, in a statistically significant manner, the best solution $x^*$ which performs better among the finite set of $k$ possibilities found during the optimization.
R\&S methods usually follow one of the following paradigms: the indifference-zone (IZ), the Bayesian, or the optimal computing budget allocation (OCBA) approaches, as described in \cite{kim2006,branke2007,hong2015}.
Unfortunately, these methods introduce more computational burden and globally optimal solutions might not be analyzed in the R\&S phase, because they are not visited during the search. Heuristic algorithms provide no optimality guarantee and, due to simulation noise, estimates of the average may be far from the real average.
As a consequence, improving solutions could be discarded and the search might never explore some portions of the search space which would further improve $x^*$.

This paper introduces a fully-sequential, efficient and robust simulation-based optimization technique based on paired t-tests and IZ selection.
Differently from current approaches, the algorithm detects significant differences by considering the relationship between probabilities $\alpha$ and $\beta$ of making an error of type I and type II. 
Given a null hypothesis $H_0$ and an alternative hypothesis $H_1$, $\alpha$ is the probability to reject $H_0$ when $H_0$ is true and $\beta$ is the probability to fail to reject $H_0$ when $H_0$ is false.
However, to compute $\beta$, a significant difference in means for which $H_0$ is assumed to be false and $H_1$ to be true has to be defined. According to observed statistical evidence, the algorithm reactively adapts the sample size of estimators when comparing pairs of different solutions during the search process. 
Also, the reactive mechanism keeps in memory all the evaluations of solutions previously visited during the search, to avoid the waste of computational budget if a configuration has to be compared multiple times.
Comparisons are done with a naive scheme which uses fixed sample sizes during the optimization, and with a R\&S procedure called Sequential Selection with Memory (SSM) which was developed by \cite{pichitlamken2006}. Results show that the technique proposed in this work is more efficient than SSM in all experiments.

The rest of the paper is structured as follows. 
Section \ref{sec:problem} defines the problem and provides the mathematical notation used in the rest of the paper. Section \ref{sec:saa} outlines the techniques which can be used to estimate the expected value of simulators, while Section \ref{indifference} overviews IZ methods.
Section \ref{reactive} presents the reactive sample size algorithm based on parametric tests and indifference-zone selection.
Finally, Section \ref{experiments} designs the experiments and analyzes the results.

\section{Problem Definition}
\label{sec:problem}
Let $F$ be a stochastic function that models a real world problem. The output of $F$ depends on some decision variables $x$ and on a random vector $\xi$ that represents the stochasticity of the problem. The expectation of $F$ is defined as
\begin{equation}\label{expectation}
f(x) = \mathbb{E} [F(x,\xi)]
\end{equation}

and it can be estimated by using a sample $\xi_1, ..., \xi_n$ of independent identically distributed (i.i.d.) realizations of the random vector $\xi$, in order to compute the Sample Average Approximation (SAA) of (\ref{expectation}) as
\begin{equation}\label{sample_average_approximation}
\hat{f_n}(x) = \frac{1}{n} \sum_{i=1}^{n} F(x,\xi_i).
\end{equation}

If the sample $\xi_1, ..., \xi_n$ is i.i.d., by the Law of Large Numbers, as $n$ approaches infinity $\hat{f_n}(x)$ converges to $f(x)$ and so $\hat{f_n}(x)$ is an unbiased estimator of $f(x)$. Moreover, if the variance of $F$ is finite, by the Central Limit Theorem $\hat{f_n}(x)$ asymptotically follows a normal distribution with mean $f(x)$ and variance $\sigma^2/n$ where $\sigma^2$ is the variance of $F$. As a consequence, the accuracy of the estimation increases with sample size $n$, but this also increments the computational burden. See also \cite{shapiro,kim2015}.
The problem is defined in the constraints-defined region $\Theta$ in which $x$ can assume values as
\begin{equation}
\min_{x \in \Theta} f(x).
\end{equation} 

The SAA defined in (\ref{sample_average_approximation}) can be used as objective function by heuristic optimization techniques, in order to optimize $f(x)$.
However, the output of simulations follows some distribution which may vary across $\Theta$.
The presence of noise might require large samples in order to obtain sufficiently accurate estimates, so comparing the performance of different configurations is not straightforward. 
A comparison can achieve a different outcome if estimates, and not the real averages, are employed. Furthermore, $\hat{f}_n(x)$ is computed by using $n$ multiple simulations which might take a considerable amount of time to run.

To obtain effective simulation-based optimization strategies, finding a tradeoff between accuracy of estimation and total running time is important, and the sample size of estimators needs to be chosen consequently.
It is not a trivial task to compare the performance of different configurations of variables. 
Given any configuration $x$, $f(x)-\hat{f_n}(x)$ defines an error $\epsilon_n(x)$ that goes to 0 only in the limit of $n$ going to infinity.
As a consequence, when comparing two configurations $x_1$ and $x_2$, the difference $\hat{f_n}(x_1) - \hat{f_n}(x_2)$ is not sufficient to decide which configuration has a better average.
In fact, consider $f(x_1) = \hat{f}_n(x_1) + \epsilon_n(x_1)$ and $f(x_2) = \hat{f}_n(x_2) + \epsilon_n(x_2)$.
The difference of the approximations can also be written as
\begin{equation}
f(x_1)-f(x_2) = \hat{f}_n(x_1) + \epsilon_n(x_1) - \hat{f}_n(x_2) - \epsilon_n(x_2).
\end{equation}
The sample size $n$ should be properly chosen in order to 
identify with a certain probability the configuration with a better average.

\section{Sample Average Approximation and Shrinking-balls}
\label{sec:saa}
In the mathematical optimization literature, there are two main approaches to compute SAA estimators: keeping the sample size fixed, or making it variable.
The simplest approach is to pick a sufficiently large sample size \textit{a priori} and to keep it fixed along the optimization process. 
To determine the fixed sample size some preliminary experiments can be run, by comparing configurations in different areas of $\Theta$ in order to determine the minimum sample size required to statistically differentiate configurations. 
See for example \cite{banks2010}.
More advanced techniques start the optimization process with a small $n$, and then they constantly increase it at a rate that asymptotically guarantees the convergence of estimators to expectations during the optimization. Examples of this type are \cite{mello}, \cite{pasupathy}, \cite{polak}. However, when comparing two configurations, such approaches do not consider the estimated improvement $\hat{f_n}(x_1) - \hat{f_n}(x_2)$ or the error $\epsilon_n(x_1) - \epsilon_n(x_2)$ in order to decide how to increase the sample size. 
These differences are taken into account in works based on adaptive samples such as \cite{bastin2004, krejic2014}. 
These techniques use ``trust region'' or ``line search'' methods to move in the neighborhood of the current configurations, according to how $f(x)$ is expected to change around the best current configuration $x_{current}$. 
As a consequence, when a new configuration $x_{new}$ is proposed, the sample size $n$ used for evaluating $\hat{f}_n(x_{new})$ is defined according to how the chosen strategy expects that $\hat{f}_n(x_{new}) - \hat{f}_n(x_{current})$ and $\epsilon_n(x_{new}) - \epsilon_n(x_{current})$ are going to change. 
This means that these techniques implicitly tie together the exploration strategy and the variation of the sample size $n$ during the optimization.

In contrast to SAA methods, an alternative approach is to approximate the value of the objective function by averaging single evaluations sampled within a hypersphere called shrinking ball, defined around the configuration to be evaluated. \cite{andradottir2010,kiatsupaibul2015} are examples of this scheme.
In shrinking-ball methods, the neighborhood of a solution $x$ is defined by a ball of a certain radius $r$. The radius of the ball can be reduced as the optimization progresses as in \cite{andradottir2010} or it can be kept constant as in \cite{kiatsupaibul2015}.
When a new single evaluation is obtained, shrinking-ball methods require to compute a distance metric such as the Euclidean distance with respect to the points which have been previously evaluated, to possibly update the respective hyperspheres. 
As stated in \cite{linz2017}, shrinking-ball methods tend to perform worse than sample average approximation methods when the amount of noise in the objective function increases.

\section{Indifference-zone Methods}
\label{indifference}
In the IZ scheme, the target is to select the best configuration $x^*$ among a finite set of $k$ configurations, where $x^*$ is better than all other configurations in the set by at least $\delta$ and the probability of correct selection (PCS) is $1-\alpha > 0$.
$\delta$ is called the IZ parameter, and it defines the minimum difference in means considered to be worth detecting.
In contrast, if the difference in means between two configurations is within $\delta$, it is indifferent to consider as better one of the two.
Also, let us define the probability of incorrect selection (PICS) of a comparison $i$  with $PICS_i$, with $1 \leq i \leq k-1$. For both PCS and PICS, if $i=k-1$, we omit the subscript and refer to the probabilities at the end of the procedure.

The IZ formulation was first introduced by \cite{bechhofer1954}.  Subsequent works include \cite{rinott1978,nelson2001}, who proposed solutions based on a two-stage approach, and \cite{paulson1964,kim2001,pichitlamken2006,hong2006,hong2007,frazier2014,fan2016}, who adopted a fully sequential approach.
As stated by \cite{kim2001}, a stage occurs whenever the simulation of a configuration is started in order to evaluate it. 
The idea of two-stage procedures is to first gather an initial sample of the configurations in the set, and then to define the sample size required to guarantee that the best configuration is selected in the second stage. 
\cite{nelson2001} discard configurations after the first stage if there is sufficient statistical evidence to do so, while \cite{rinott1978} does not have any elimination step and consequently performs worse.
As stated by \cite{dudewicz1975}, single-stage IZ procedures cannot guarantee to find $x^*$ among a set of $k$ alternatives when the variances of the configurations are unknown. In contrast, as \cite{nelson2001} assert, two-stage IZ procedures can guarantee to find $x^*$ whenever the best configuration is at least $\delta$ better than the other configurations.

Differently from two-stage procedures, fully sequential methods sequentially compute single observations of the configurations and eliminate statistically inferior configurations. In fact, as stated by \cite{kim2001}, the goal of such procedures is to discard configurations as soon as possible while guaranteeing the PCS, to reduce the computational burden required to find $x^*$. 
But, as highlighted by \cite{frazier2014}, since most fully sequential procedures use the Bonferroni inequality to guarantee the PCS, they tend to be too conservative.
In fact, many techniques based on the IZ formulation aim at guaranteeing that
\begin{equation}
PCS \geq 1 - \sum_{i=1}^{k-1} PICS_i,
\end{equation}
where the PICS of the whole procedure is bound, according to the Bonferroni inequality, by the PICS of the $k-1$ systems in the set:
\begin{equation}
PICS \leq \sum_{i=1}^{k-1} PICS_i.
\end{equation}
Examples of these procedures, which consider the lower bound of the Bonferroni inequality to guarantee the PCS,
are \cite{kim2001,pichitlamken2006,hong2006,hong2007}.

An exception is given by \cite{frazier2014}, who proposes a Bayes-inspired indifference-zone (BIZ) procedure based on the lower bound defined by a Bayesian approach, which can also achieve a pre-specified PCS. Such a procedure is more efficient than other fully-sequential methods based on the Bonferroni inequality, in particular when the number of alternatives is very large.
Another possibility is given by \cite{fan2016}, who propose an IZ-free formulation that selects the best alternative with a user-specified PCS. 
The idea is to build a continuation region where, with a specified PCS, no elimination is made if the difference in means between two alternatives is zero, while a configuration is discarded otherwise.
Consequently, the means of any pair of configurations can be arbitrarily close as long as it is not zero, and it is not necessary to specify $\delta$.

\subsection{Indifference Zone Methods for Heuristic Search}

The R\&S methods enumerated in Section \ref{indifference} assume that no solutions have already been sampled before the beginning of the procedure. Also, they do not exploit samples obtained previously, and already visited solutions are always evaluated anew.
In contrast, during the heuristic search phase, using information about already visited solutions is desirable.
But extending techniques like \cite{kim2001,hong2006}, in order to exploit previous samples (and proving their validity), is not straightforward \cite{kim2007}.
Differently from previously mentioned R\&S approaches, \cite{pichitlamken2006,hong2007} propose methods to be applied in the heuristic search phase. 
Both works are based on the results of \cite{fabian1974,hartmann1991}, who improved the PCS bounds for the procedures proposed by \cite{paulson1964}. These approaches use a fixed triangular region, also called continuation region, to decide the sample size which should be used for each comparison in order to guarantee the PCS of the procedure.

SSM, proposed by \cite{pichitlamken2006}, is a fully-sequential scheme with elimination which keeps in memory samples of solutions found during the search. 
The samples are reused during the optimization, if previous solutions have not already been considered as statistically inferior with respect to the best. However, this procedure makes no use of common random numbers (CRN).
At each iteration of SSM an additional observation is taken for each surviving solution, which might be eliminated if the cumulative sum between the evaluations of the solution and the best falls out from the continuation region.

\cite{hong2007} propose fully-sequential R\&S methods for problems in which solutions are generated sequentially.
They propose single-elimination approaches, where each solution is considered only once, and stop-and-go schemes where each solution is considered throughout the whole optimization.
However, these methods also aim at providing an overall statistical guarantee at the end of the optimization. 
In order to fulfill this requirement using the triangular continuation region, a large number of evaluations is necessary. Therefore, these procedures are only applicable to optimization problems where the number of solutions is very small.

\section{The Reactive Sample Size Algorithm}
\label{reactive}

The algorithm proposed in this work is defined as reactive because, as defined by \cite{battiti2008}, it reacts autonomously to what happens during the optimization. The algorithm adapts the sample size $n$ used to compute each $\hat{f}_n(x)$, and it can be used as an evaluation scheme for comparing configurations during any heuristic optimization process.

At each step of the optimization, a new configuration $x_{new}$ is proposed by a heuristic algorithm and must be compared against the current best configuration $x_{current}$.
For each comparison, the algorithm reactively changes the sample size $n$ used to evaluate $\hat{f}_n(x_{new})$ and $\hat{f}_n(x_{current})$, according to observed statistical evidence.
In fact, to be confident about the outcome of comparisons, the presence of a statistically significant difference between $\hat{f}_n(x_{new})$ and $\hat{f}_n(x_{current})$ should be tested.
Precisely, it is not necessary to detect a difference between the estimators, which would require a two-sided test. 
In contrast, using a one-sided test to assess if $\hat{f}_n(x_{new}) > \hat{f}_n(x_{current})$ or $\hat{f}_n(x_{new}) < \hat{f}_n(x_{current})$ is sufficient, depending on whether one is  maximizing or minimizing the objective function. 
Upper-tailed tests are used in the case of maximization and lower-tailed tests for minimization.
The following definitions consider a maximization problem, where null and alternative hypothesis of the upper-tailed test on the difference in means are defined as
\begin{equation}
\begin{split}
H_0: \hat{f}_n(x_{new}) - \hat{f}_n(x_{current}) \leq 0 \\
H_1: \hat{f}_n(x_{new}) - \hat{f}_n(x_{current}) > 0. 
\end{split}
\end{equation}

To statistically determine if $H_1$ is true, paired or unpaired evaluations can be used to compute a statistic. 
The reactive algorithm uses paired evaluations to compute the paired t-test statistic, obtained by evaluating $x_{new}$ and $x_{current}$ on the same seeds $\xi_i$ and therefore using CRN during the optimization. 
As observed by \cite{jian2015}, using the same seeds helps to reduce the effect of noise.
The correlation among pairs of evaluations reduces the variance with respect to an unpaired statistic.
Also, the algorithm assumes that F is normally distributed and that its variance is finite.
Using the paired t-test, since the mean of paired differences corresponds to the difference of means, the statistic to test  is
\begin{equation} \label{t-statistic}
T_{n-1} = \frac{\delta_{observed}-\delta_{H_0}}{s/\sqrt{n}},
\end{equation}

where observed difference $\delta_{observed} = \hat{f}_n(x_{new}) - \hat{f}_n(x_{current})$, null hypothesis difference $\delta_{H_0}=f(x_{new})-f(x_{current})=0$ and $s$ is the sample standard deviation of paired evaluations plus a very small constant, added to avoid the possibility of dividing by zero. 
Furthermore, since $\sigma$ of F is approximated by using $s$, $T_{n-1}$ follows a t-distribution with $n-1$ degrees of freedom (d.o.f.) and the algorithm requires $n \geq 2$. 
The statistic in equation (\ref{t-statistic}) can be used to compute the p-value, which defines how unlikely it is to observe a statistic such as $T_{n-1}$ if $H_0$ is true. If this is too unlikely (threshold defined by the user-defined probability $\alpha$ of making an error of type I), then $H_0$ should be rejected.
However, it is also important to consider the probability $\beta$ to make an error of type II, so to fail to reject $H_0$ when $H_0$ is false. 
To compute $\beta$, a specific value for which $H_1$ is assumed to be true has to be defined, because $\beta$ depends on the difference $\delta_{observed}$ for which $H_0$ is assumed to be false and $H_1$ to be true. In fact, assuming that $H_1$ is true, if $T_{n-1}$ falls in the acceptance region of $H_0$ then it is unlikely that $H_0$ is false. 
In the reactive algorithm, the value for which $H_1$ is assumed to be true is $\delta_{observed}$.
When $H_1$ is true, $T_{n-1}$ follows a noncentral t-distribution with $n-1$ d.o.f. Unfortunately the noncentral t-distribution has a complex density function, but \cite{dupont1990} propose a method to approximate it using the t-distribution. Thus, in the one-tailed case, $\beta$ can be approximated as
\begin{equation} \label{beta}
\beta = 1 - T_{n-1}(\delta_{norm} \sqrt{n} - t_{n-1,\alpha}) + T_{n-1}(- \delta_{norm} \sqrt{n} - t_{n-1,\alpha})
\end{equation}

where $T_{n-1}$ is the cumulative distribution function of the t-distribution with $n-1$ d.o.f., $\delta_{norm} = \delta_{observed} / s$ and $t_{n-1,\alpha}$ is the quantile $x$ of the t-distribution with $n-1$ d.o.f. such that $T_{n-1}(x)=\alpha$.
As a consequence, given a paired sample, equation (\ref{beta}) can be used to iteratively find the minimum sample size $n$ that should be used to test a one-tailed hypothesis with error probabilities $\alpha$ and $\beta$ as
\begin{equation} \label{samplesize}
n = \frac{(t_{n-1,\alpha} + t_{n-1,\beta})^2}{\delta_{norm}^2}.
\end{equation}

However, in real world problems, one might not be interested to correctly detect very small differences between means. Or, from a different perspective, it might be too expensive to statistically differentiate between two means that are very similar. 
Precisely, if $x_{new}$ and $x_{current}$ have a very similar performance and so $\delta_{observed}$ is smaller than a certain user-defined $\delta$, the comparison should be done heuristically by using only the values of $\hat{f}_n(x_{new})$ and $\hat{f}_n(x_{current})$. 
The value of $\delta$ is expressed as a percentage of $\hat{f}_n(x_{current})$, because in many cases the user does not know \textit{a priori} the best possible result which can be obtained by the optimization.
Furthermore, because of the stochasticity of the objective function, the algorithm also considers the impact that observed standard deviation $s$ has on the estimation of the minimum sample size $n$ required to reject $H_0$.
In fact, high levels of noise might require a huge amount of evaluations in order to statistically differentiate between $\hat{f}_n(x_{new})$ and $\hat{f}_n(x_{current})$ by at least $\delta$.
So, before checking if a comparison should be done heuristically, $\delta$ is normalized by $s$. 
As a consequence, to apply the heuristic solution, the algorithm checks if
\begin{equation} 
\delta_{norm} < \frac{\delta}{s}.
\end{equation}

In the heuristic solution, $\hat{f}_n(x_{new})$ is evaluated at least as many times and on the same set of seeds as $\hat{f}_n(x_{current})$.
In fact, during the optimization process the reactive scheme keeps track of the sample size $n_{current}$ used to evaluate $\hat{f}_n(x_{new})$ when $\delta_{norm} < \delta/s$, and any sample size which lowers the current value of $n_{current}$ is not accepted. As a consequence, the sample size used in this case can only increase as the optimization advances.
As suggested in \cite{hong2009}, the sample size should increase as the optimization proceeeds towards a local minimum, because it is more difficult to detect the difference between solutions which tend to have similar values. 
When the current evaluation is far from any minimum present in $\Theta$, a small sample size should be sufficient to distinguish the performance of different configurations. As the process goes towards locally optimal points, comparing diverse solutions becomes harder and so additional evaluations are necessary.

\subsection{Parameters of the algorithm} 
The main parameters of the algorithm are three: the required probability $\alpha_{req}$ of making an error of type I, the required probability $\beta_{req}$ of making an error of type II and the minimum required difference $\delta$ between averages to apply a statistical test.
To provide additional flexibility to the algorithm, a minimum number $n_{min}$ and a maximum number $n_{max}$ of function evaluations can be set for each configuration in comparisons.
If during a comparison $n_{max}$ is reached, but $\alpha_{req}$ and $\beta_{req}$ have not yet been satisfied, a decision is taken by considering only the values of $\hat{f}_n(x_{new})$ and $\hat{f}_n(x_{current})$. 
In the experiments $n_{min} = 2$ and $n_{max} = \infty$, so the algorithm autonomously decides when to stop. 

\subsection{Statistical guard}\label{guard}
When increasing $n$ using (\ref{samplesize}), the update is not done in one-shot but iteratively, updating the pair $\hat{f}_n(x_{new})$ and $\hat{f}_n(x_{current})$ with one evaluation at the time. 
This is done because (\ref{samplesize}) defines an estimation of the minimum sample size required to test the hypothesis, which is based on the observed sample. But the observed sample is not always representative of the whole population, and so estimations might not be correct. 
As $n$ increases, the sample is going to be more and more representative; but by checking iteratively p-value and $\beta$ a lot of unnecessary evaluations are saved. 
In fact, at each update, p-value and $\beta$ are compared with $\alpha_{req}$ and $\beta_{req}$; if such requirements are satisfied, then a statistically significant decision can be taken. 
The same is done when $\delta_{norm} < \delta/s$ and the heuristic solution is applied, because during the optimization the sample size $n$ used in the heuristic case might become too large for certain comparisons. 
In fact, by computing iteratively p-value and $\beta$ even in the heuristic solution, the statistical guard can check if sufficient statistical evidence has been observed in order to take a statistically significant decision.
The statistical guard is defined in Algorithm \ref{statistical_comparison}.
\subsection{Outline of the algorithm}
The reactive sample size algorithm is outlined in Algorithm \ref{reactive_algorithm}. At each step of the optimization process, given two configurations $x_{new}$ and $x_{current}$, the algorithm first evaluates both configurations on $n_{min}$ seeds in order to obtain $\hat{f}_n(x_{new})$ and $\hat{f}_n(x_{current})$, where $n = n_{min}$. 
Then, p-value and $\beta$ are computed using (\ref{t-statistic}) and (\ref{beta}), respectively. If such values satisfy the probabilities of making an error of type I and II required by the user, then a decision which is considered to be statistically significant can be taken; otherwise, the sample size needs to be increased according to (\ref{samplesize}). 
In both cases, if $\delta_{norm} < \delta/s$, the decision is taken heuristically by considering only the values of $\hat{f}_n(x_{new})$ and $\hat{f}_n(x_{current})$. The optimization continues as long as there is a sufficient amount of budget left, where the budget is defined as number of objective function evaluations.
Algorithm \ref{reactive_algorithm} is built on the assumption that it is called by some heuristic optimization process when some $x_{new}$ has to be compared against $x_{current}$, with $x_{new} \neq x_{current}$.
In Algorithm \ref{reactive_algorithm}, on line \ref{update_current} a new evaluation $F(x_{current},\xi_i)$ is computed only if $\hat{f}_i(x_{current})$ has not been evaluated on $\xi_i$ yet.
On line \ref{line_underestimate}, $n$ is computed again in the case that the initial sample produces an underestimate of the minimum sample size required to statistically differentiate between the performance of $x_{new}$ and $x_{current}$.

\begin{algorithm}[h]
\caption{Statistical guard used to check if a statistically significant decision can be taken}\label{statistical_comparison}
\begin{algorithmic}[1]
\Procedure{statisticalGuard}{$x_{new}, x_{current},\alpha_{req},\beta_{req},\delta,n,n_{current}$}
\State Compute $\delta_{observed}, s, \delta_{norm}$
\State p-value $\leftarrow 1 - T_{n-1}(\frac{\delta_{observed}}{s/\sqrt{n}})$ \Comment{In a minimization problem: p-value $\leftarrow T_{n-1}(\frac{\delta_{observed}}{s/\sqrt{n}})$}
\State $\beta \leftarrow 1 - T_{n-1}(\delta_{norm} \sqrt{n} - t_{n-1,\alpha}) + T_{n-1}(- \delta_{norm} \sqrt{n} - t_{n-1,\alpha})$
\If{$\beta \leq \beta_{req}$}
	\If{p-value $ \leq \alpha_{req}$}
		\State $x_{current} \leftarrow x_{new}$
		\State $n_{current} \leftarrow max(n,n_{current})$
	\EndIf
	\State return
\EndIf
\EndProcedure
\end{algorithmic}
\end{algorithm}
\begin{algorithm}[H]
\caption{Reactive sample size algorithm}\label{reactive_algorithm}
\begin{algorithmic}[1]
\Procedure{reactiveComparison}{$x_{new}, x_{current},\alpha_{req},\beta_{req},\delta,n_{min},n_{max},n_{current}$}
\State Compute $\hat{f}_n(x_{new})$ using the same sample $\xi_1,...,\xi_n$ as $\hat{f}_n(x_{current})$, where $n = n_{min}$
\State statisticalGuard($x_{new}, x_{current},\alpha_{req},\beta_{req},\delta,n_{min},n_{current}$)
\State $n \leftarrow (t_{n-1,\alpha} + t_{n-1,\beta})^2 / \delta_{norm}^2.$
\State $i \leftarrow n_{min}$
\While{$i \leq n$}
	\State Update $\hat{f}_{i}(x_{new})$ and $\hat{f}_{i}(x_{current})$ using respectively  $F(x_{new},\xi_{i+1})$ and $F(x_{current},\xi_{i+1})$ \label{update_current}
	\State statisticalGuard($x_{new}, x_{current},\alpha_{req},\beta_{req},\delta,i,n_{current}$)
	\If{$\delta_{norm} \leq \delta/s$}
		\State $k \leftarrow i$
		\While{$k \leq n_{current}$}
			\State Update $\hat{f}_{k}(x_{new})$ using $F(x_{new},\xi_{k+1})$			
			\State statisticalGuard($x_{new}, x_{current},\alpha_{req},\beta_{req},\delta,k,n_{current}$)					
			\State $k \leftarrow k + 1$
		\EndWhile
		\State heuristicDecision($x_{new},x_{current},n_{current}$)
	\EndIf		
	\State $i \leftarrow i +1$
	
	\If{$i = n_{max}$}
		\State heuristicDecision($x_{new},x_{current},n_{max}$)
	\EndIf	
	
	\If{$i = n+1$}
		\State $n \leftarrow (t_{n-1,\alpha} + t_{n-1,\beta})^2 / \delta_{norm}^2.$ \label{line_underestimate}
	\EndIf
	\EndWhile
\EndProcedure
\end{algorithmic}
\end{algorithm} 

\section{Numerical Experiments}
\label{experiments}
The reactive algorithm is tested on four objective functions, by comparing its performance with SSM and a more naive scheme which uses a fixed sample size.
The first set of experiments considers three deterministic functions extended with different levels of normally distributed noise.
The fourth objective function used in the tests is Hotelsimu, a real simulation-based optimization tool for hotel revenue management proposed by \cite{hotelsimu}. 
\begin{algorithm}[t]
\caption{Heuristic decision taken by considering only the values of estimators}\label{heuristic_decision}
\begin{algorithmic}[1]
\Procedure{heuristicDecision}{$x_{new},x_{current},n$}
		\If{$\hat{f}_{n}(x_{new}) > \hat{f}_{n}(x_{current})$}	\Comment{In a minimization problem: $\hat{f}_{n}(x_{new}) < \hat{f}_{n}(x_{current})$}
			\State $x_{current} \leftarrow x_{new}$
		\EndIf							
		\State return
\EndProcedure
\end{algorithmic}
\end{algorithm}

\subsection{Functions}
The reactive algorithm is evaluated on  Rastrigin, Griewank and Rosenbrock functions, which are extended by adding 1\%, 5\%, 10\% and 20\% of their output as normally distributed noise. Rastrigin function is defined as
\begin{equation}
f(x) = 10 \; d + \sum_{i=1}^d x_i^2 - 10 cos(2 \pi x_i) \nonumber,
\end{equation}

Griewank function as
\begin{equation}
f(x) = \sum_{i=1}^d \frac{x_i^2}{4000} - \prod_{i=1}^d cos(\frac{x_i}{\sqrt{i}}) + 1, \nonumber
\end{equation}

and Rosenbrock function as
\begin{equation}
f(x) = \sum_{i=1}^{d-1} 100 \; (x_{i+1} - x_i^2)^2 + (1 - x_i)^2 \nonumber,
\end{equation}

where d is the number of dimensions. In the tests, $d=10$. As proposed in various works such as \cite{yang2010}, experiments consider Rastrigin function in $[-5.12,5.12]^d$, Griewank function in $[-600,600]^d$ and Rosenbrock function in $[-5,5]^d$.

\subsection{HotelSimu}
HotelSimu is a simulation-based optimization approach based on dynamic pricing for hotel revenue management, proposed by \cite{hotelsimu}. 
It simulates the hotel booking scenario, from reservations with different characteristics (i.e. time to arrival, length of stay, group size) to possible cancellations and walk-ins, in order to maximize the total revenue of a hotel over a simulated period of time.
This simulation-based optimization approach takes as input data daily statistics such as expected number of reservations and cancellations, expected length of stay per reservation and expected number of rooms per reservation.
Arrivals and cancellations curves are defined by a set of parametric models based on the RIM quantifiers defined by \cite{yager1998}, which are commonly used in decision making.
Given an expected number of reservations or cancellations for an arrival day, such parametric models distribute the events across the booking horizon so that on average their sum corresponds to the number of events set for the last day.
For each day, events are assumed to follow a non-homogeneous Poisson process with an expected value computed by the parametric model.

The simulation model is composed by several modules: an event generator, a hotel registry, a dynamic pricing model and a price elasticity model. The event generator creates interspersed reservations and cancellations, the hotel registry contains the state of the hotel, the dynamic pricing model computes the price of each reservation and the price elasticity model simulates how customers accept or discard reservation offers.
As long as there still is budget left, the heuristic optimization strategy iteratively proposes configurations to the simulator.
The configurations proposed by the optimizer are the 6 continuous parameters of the 4 linear multipliers of the dynamic pricing model proposed by \cite{bayoumi2013dynamic}, which define the slopes of the linear functions and consequently how reservations are priced. Then, for each reservation request, the multipliers propose a price obtained by multiplying an average reference price $P_{ref}$ by the value of each multiplier, which can take values in $[0.60,1.40]$.

In the simulator, for each simulated reservation day, a random sequence of reservation requests and cancellations is generated. For each reservation request created by the event generator, the dynamic price model assigns a price which depends on its features, on the remaining capacity of the hotel and on the parameters proposed at each step by the heuristic optimization.
As a result, reservation requests become reservation offers. These offers are fed to the price elasticity model, which simulates the decision of a customer regarding the reservation offer which has been received. If the customer accepts the offer, the remaining capacity of the hotel is decreased and the hotel registry is updated. Also, if the accepted offer is not canceled before the end of the simulation period, it is considered in the evaluation of the total revenue to be passed to the optimizer.
However, as previously mentioned, the output of the simulation is stochastic and so the evaluation of each configuration is estimated as the SAA of the total revenue of $n$ simulation runs.

\subsection{Setup}
The experiments adopt a simple version of random local search (RLS) as optimization strategy, to easily observe the impact of different comparison policies on the optimization. 
Although more advanced heuristic algorithms could furtherly improve the results, these techniques would also introduce additional complexities into the analysis.
Such procedures, which can also benefit by the reactive scheme proposed in this work, include methods like evolutionary algorithms, adaptive random search and Bayesian Optimization.

In RLS, a new candidate solution $x_{new}$ is sampled from an interval defined around the current best solution $x_{current}$, according to some distribution. 
In this work, a uniform distribution is used to sample new solutions. A step size is used to define, as a percentage of the interval in which the function is defined, the boundaries of the interval around the best current solution in which new candidate solutions are sampled.
Consequently, diverse step sizes correspond to search policies with different levels of locality.
A step size of 1 makes the search global, and the optimization corresponds to pure random search. 

RLS is employed in the experiments with static step size $\in \lbrace 0.2, 0.4, 1 \rbrace$, to analyze how the comparison schemes deal with estimators which are on average more or less different.
Each optimization strategy is compared by using a fixed sample size policy, the reactive sample size scheme or SSM to compare solutions during the search.
When using fixed sample size, SAAs are computed using sample sizes $\in \{1,2,20\}$ throughout the whole optimization process. 
In the reactive sample size case, the parameters of the algorithm are $\alpha_{req} = 0.1$, $\beta_{req} = 0.4$ and $\delta = 0.01$. 
The value of $\beta_{req}$ has been set after having performed various experiments and observed this value leads on average to efficient optimization runs, while mantaining robustness.
However, if the value of $\alpha_{req}$ is given, then $\beta_{req}$ can be modified in order to change the robustness of the optimization process.
For SSM, $\delta$ and $\alpha_{req}$ are set to the same values as in the reactive scheme, and in both algorithms the minimum sample size $n_0$ is set to 2. Also, as suggested in \cite{pichitlamken2006}, $c = 1$; for more details about this parameter, please refer to the original paper of SSM.

\begin{figure}
\centering
\begin{minipage}{.5\textwidth}
  \centering
  \includegraphics[height=170pt,width=190pt]{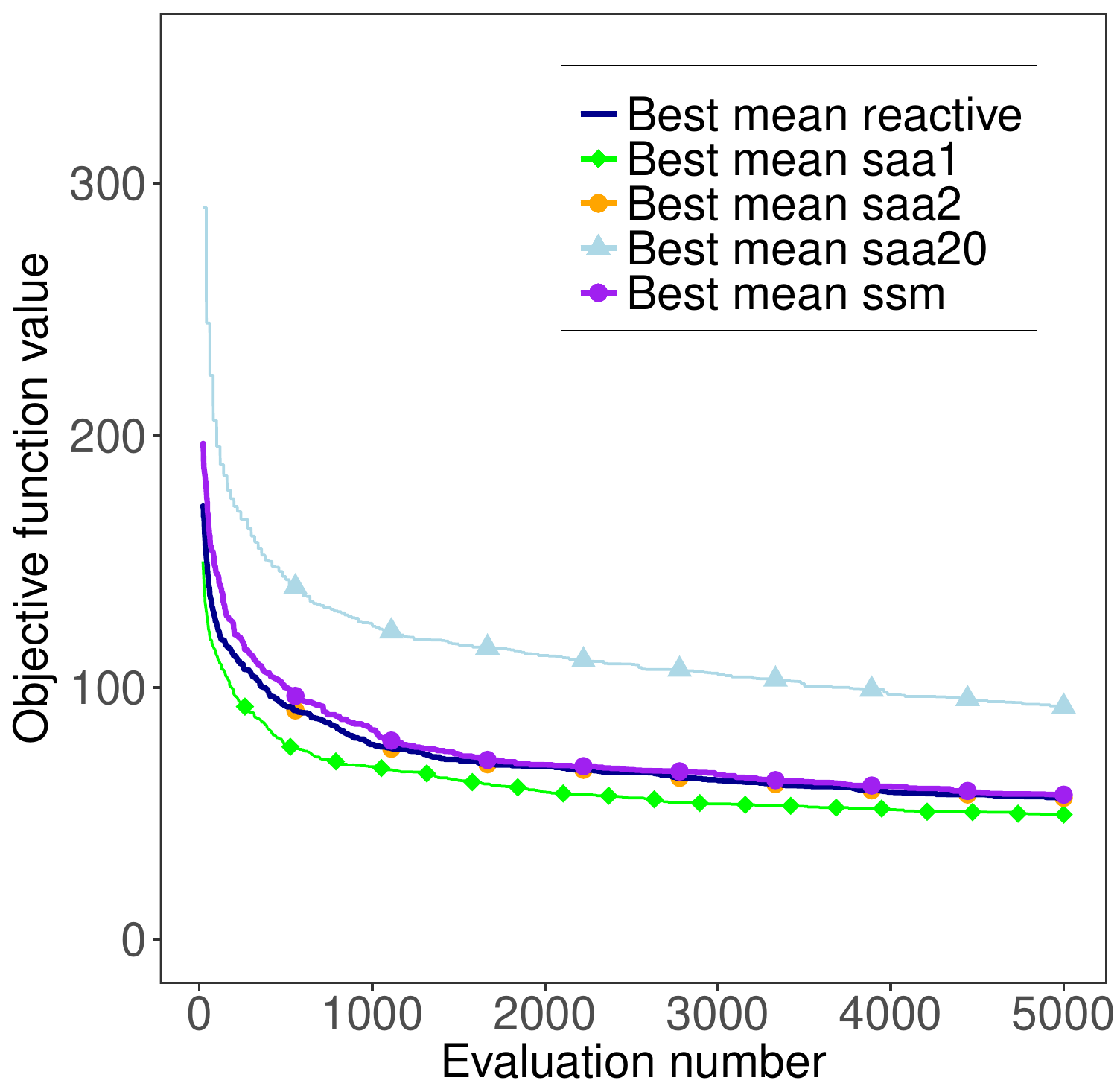}
  \caption{Griewank, step = 1.0 and noise = 1\%.}
  \label{fig:test7}
\end{minipage}%
\begin{minipage}{.5\textwidth}
  \centering
  \includegraphics[height=170pt,width=190pt]{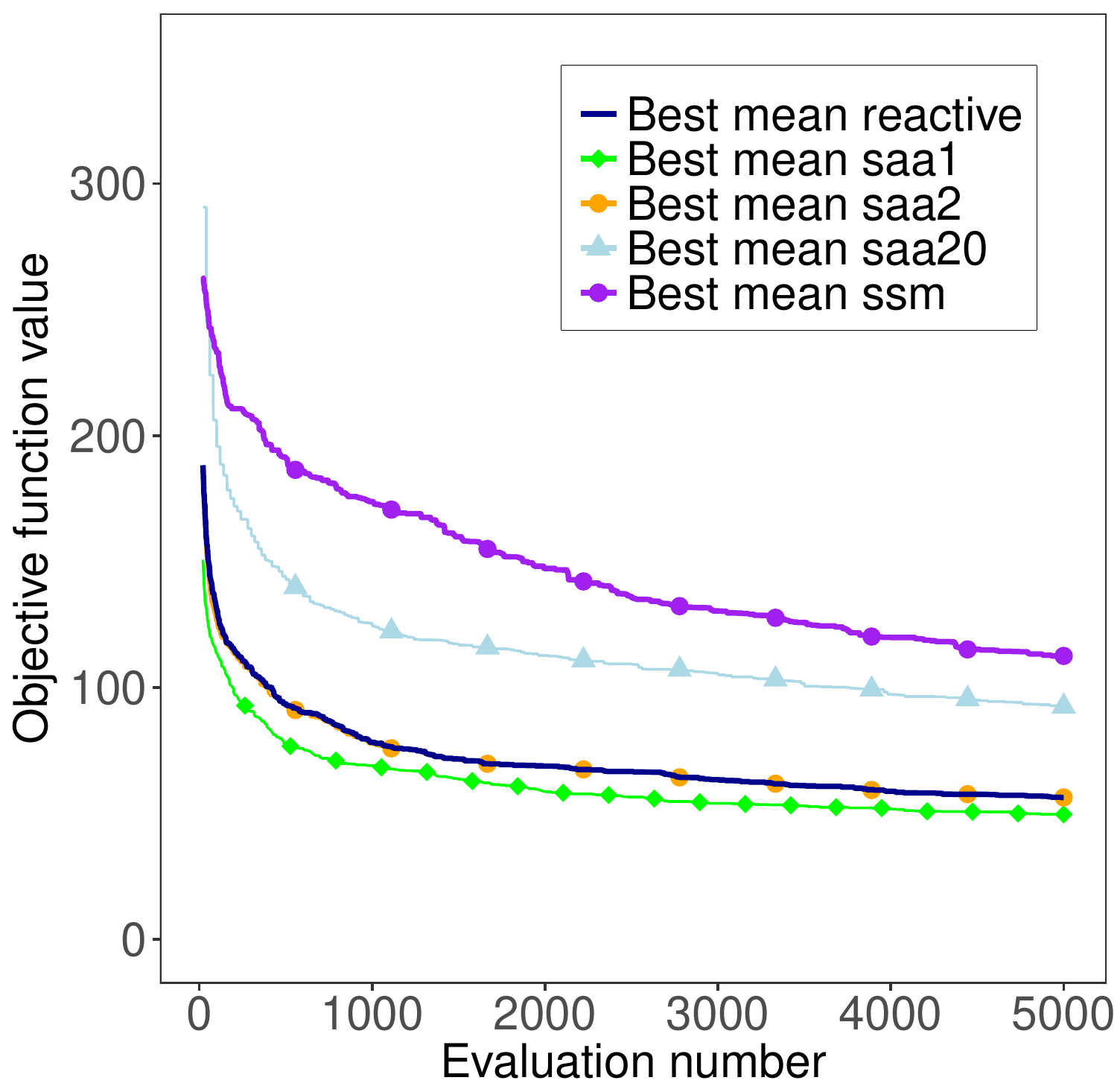}
  \caption{Griewank, step = 1.0 and noise = 5\%.}
  \label{fig:test8}
\end{minipage}
\end{figure}
\begin{figure}
\centering
\begin{minipage}{.5\textwidth}
  \centering
  \includegraphics[height=170pt,width=190pt]{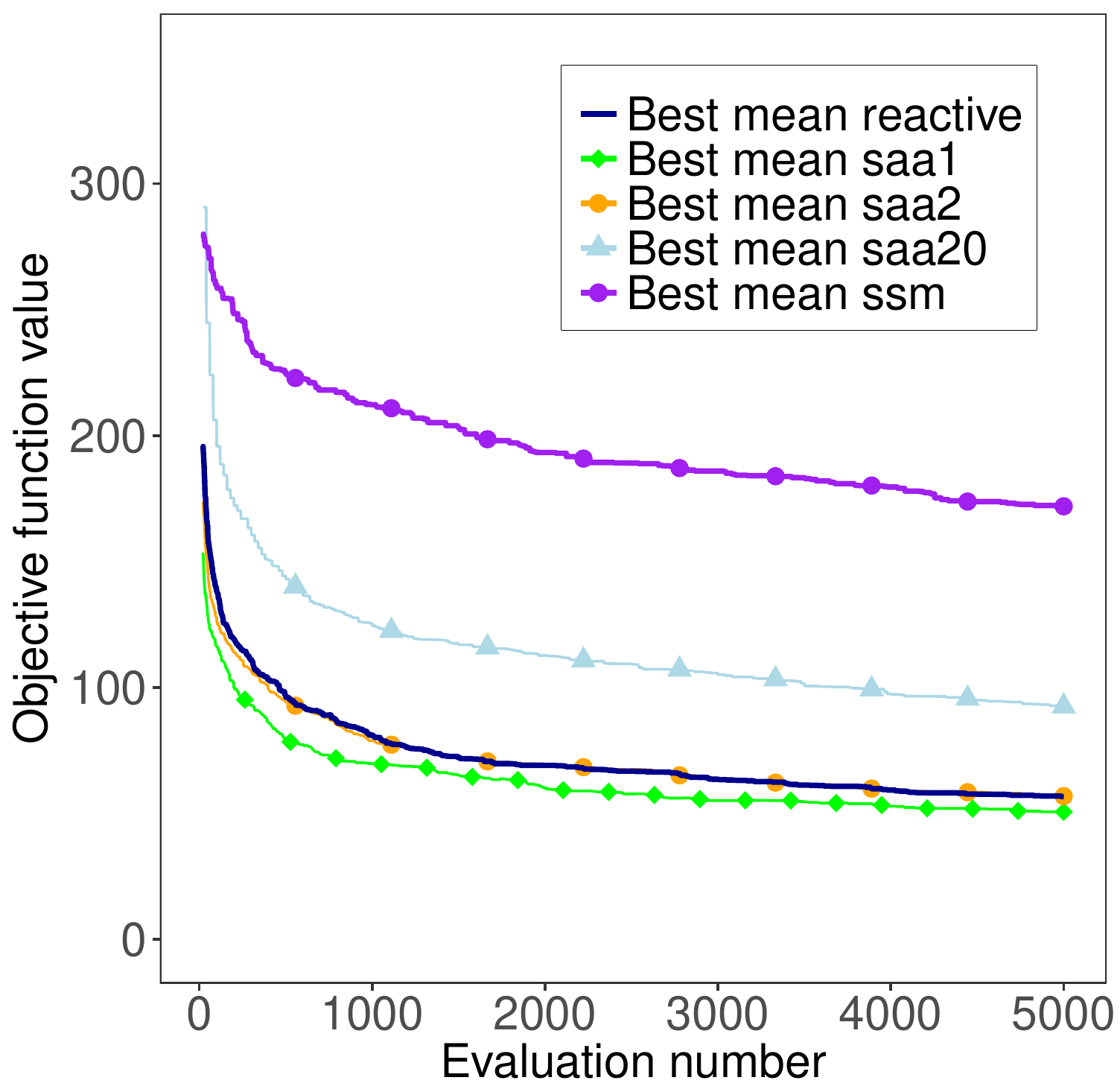}
  \caption{Griewank, step = 1.0 and noise = 10\%.}
  \label{fig:test9}
\end{minipage}%
\begin{minipage}{.5\textwidth}
  \centering
  \includegraphics[height=170pt,width=190pt]{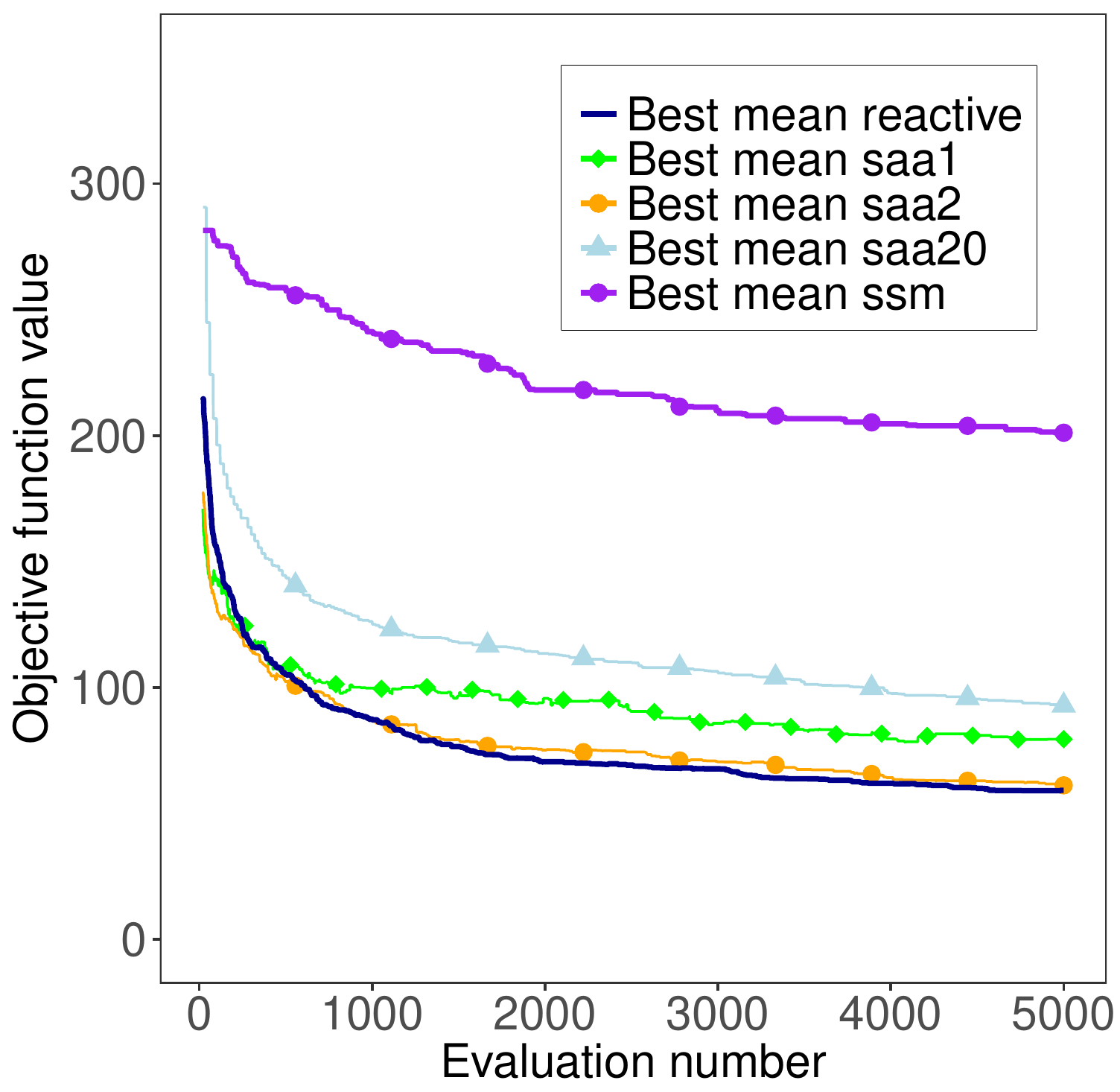}
  \caption{Griewank, step = 1.0 and noise = 20\%.}
  \label{fig:test10}
\end{minipage}%
\end{figure}

\subsection{Results}
In the first set of experiments, Griewank, Rastrigin and Rosenbrock functions are extended with artificial levels of noise and minimized. 
Each experiment is based on 100 macroreplications, and each optimization process uses budget = 5000.
Lines in the figures represent the mean noiseless value of the best configuration found during the optimization. 
Similarly to \cite{pichitlamken2006}, the first part of the analysis presents results which employ only pure random search as search policy.
In contrast, the second part extends the study by showing also the results obtained by using more local versions of RLS.

\begin{figure}
\centering
\begin{minipage}{.5\textwidth}
  \centering
  \includegraphics[height=170pt,width=190pt]{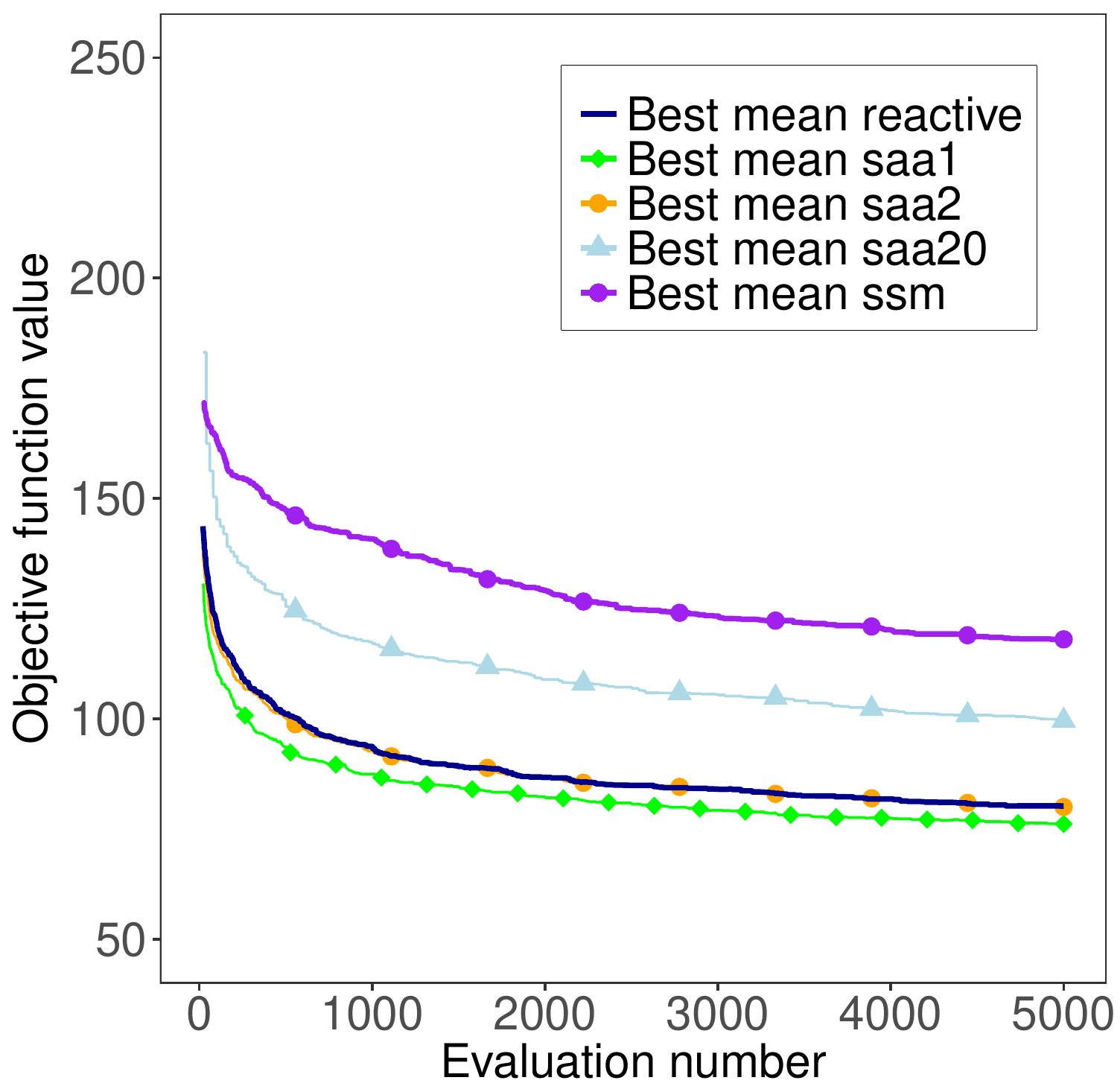}
  \caption{Rastrigin, step = 1.0 and noise = 5\%.}
  \label{fig:test11}
\end{minipage}%
\begin{minipage}{.5\textwidth}
  \centering
  \includegraphics[height=170pt,width=190pt]{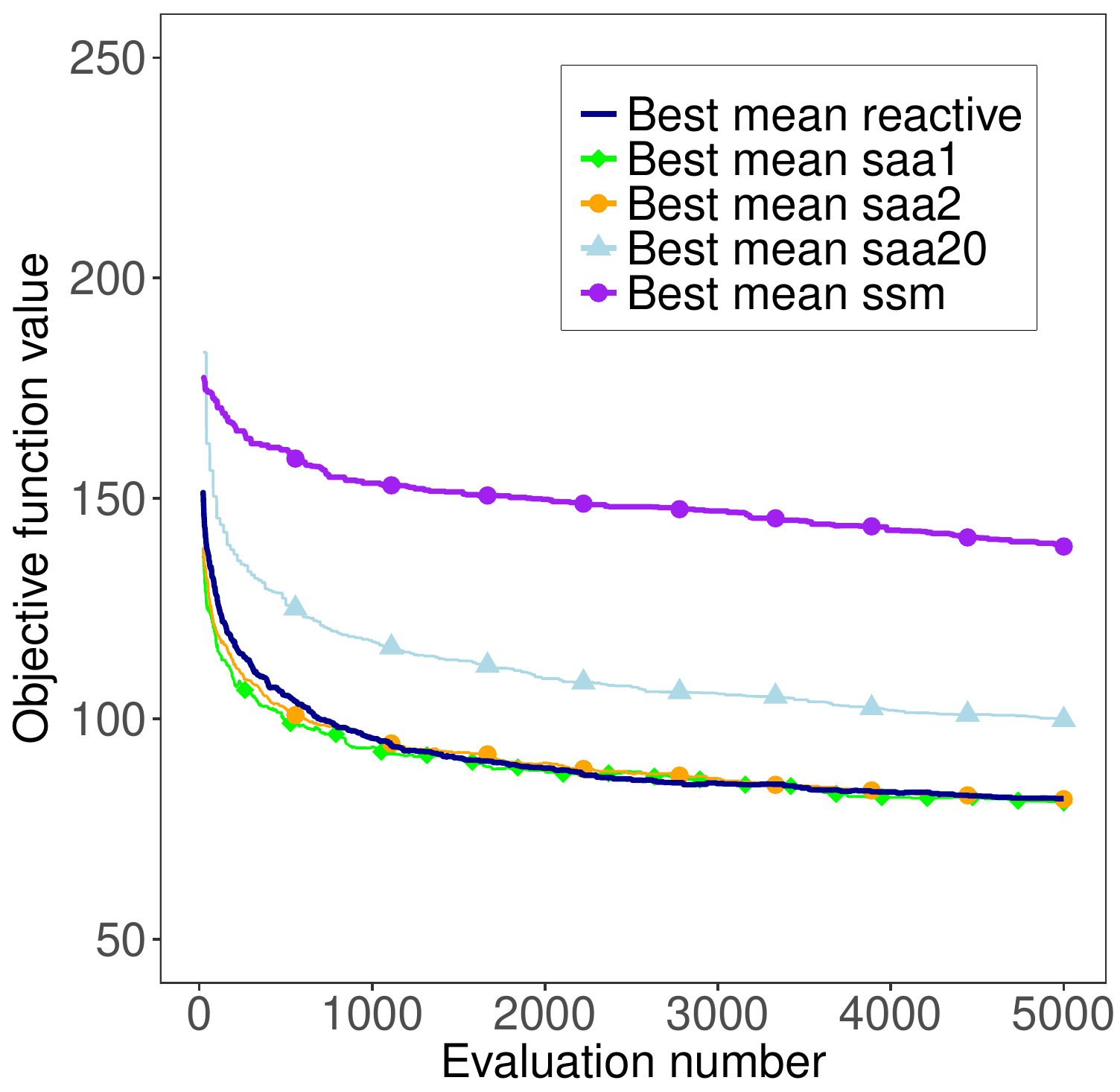}
  \caption{Rastrigin, step = 1.0 and noise = 10\%.}
  \label{fig:test12}
\end{minipage}
\end{figure}

\begin{figure}
\centering
\begin{minipage}{.5\textwidth}
  \centering
  \includegraphics[height=170pt,width=190pt]{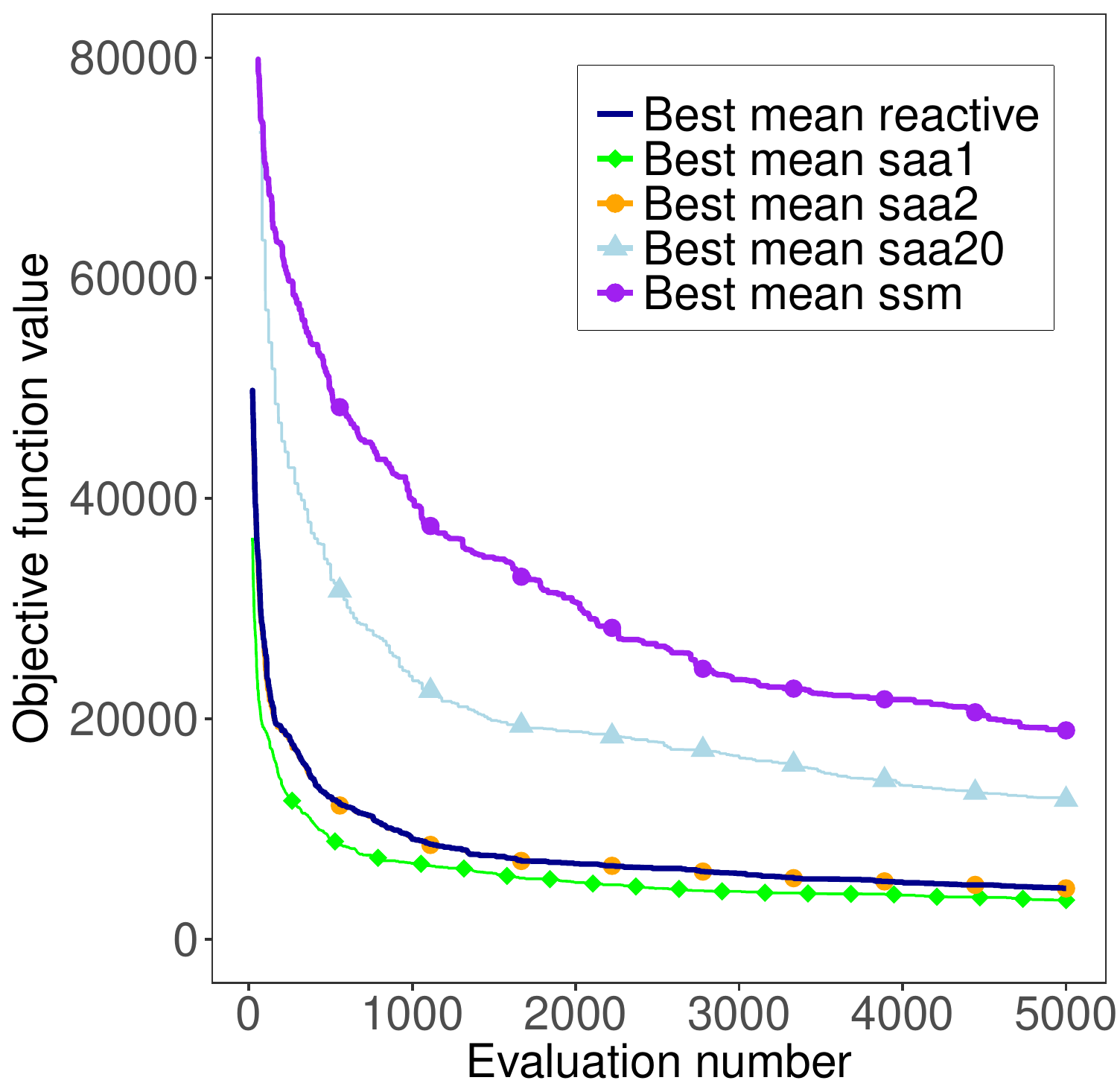}
  \caption{Rosenbrock, step = 1.0 and noise = 5\%.}
  \label{fig:test13}
\end{minipage}%
\begin{minipage}{.5\textwidth}
  \centering
  \includegraphics[height=170pt,width=190pt]{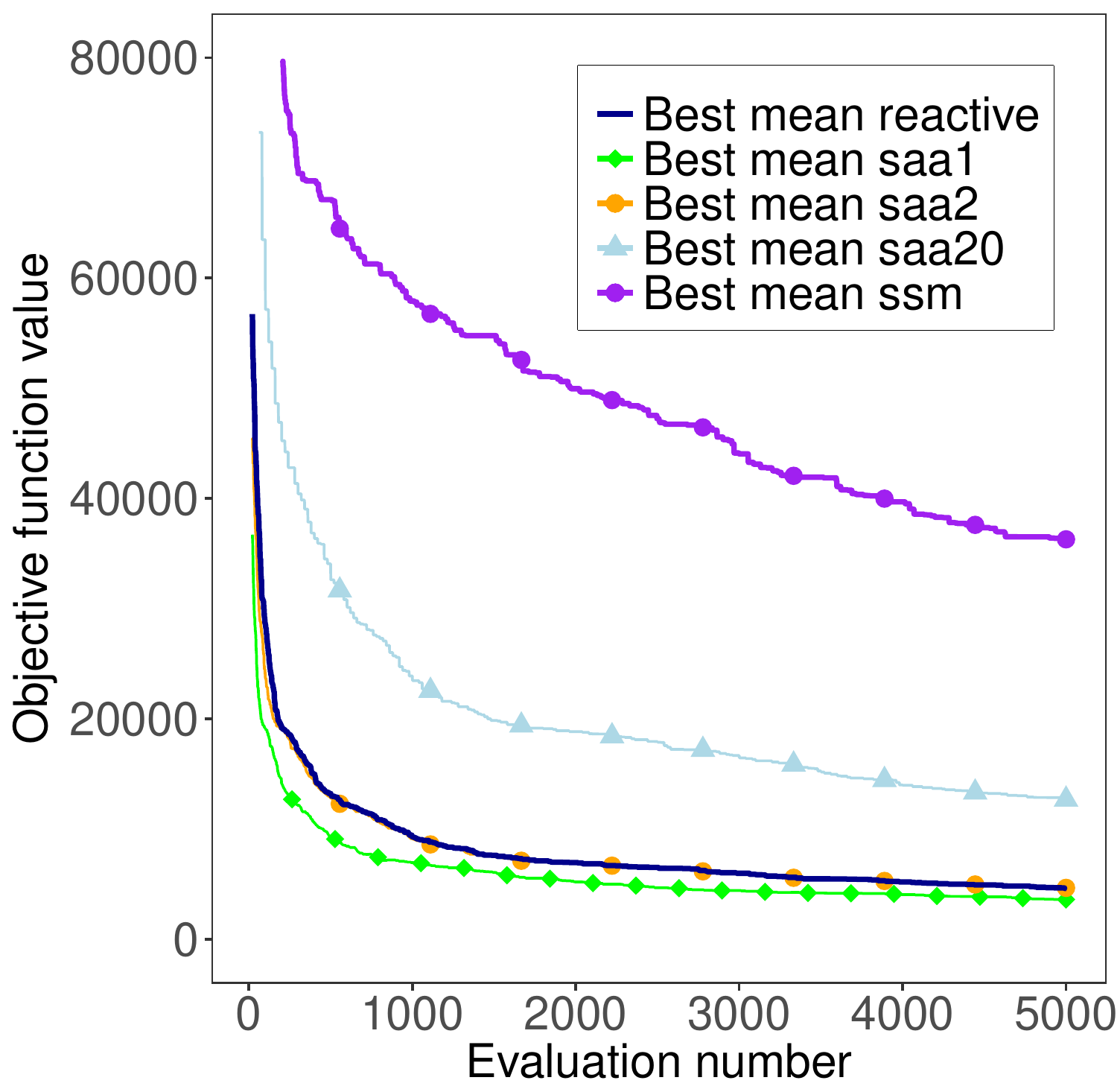}
  \caption{Rosenbrock, step = 1.0 and noise = 10\%.}
  \label{fig:test14}
\end{minipage}
\end{figure}

Figure \ref{fig:test7}, Figure \ref{fig:test8}, Figure \ref{fig:test9} and Figure \ref{fig:test10} show optimization results when Griewank function contains respectively 1\%, 5\%, 10\% and 20\% of its output as normally distributed noise. 
These figures show scenarios in which using a fixed sample size of 20 does not bring any benefit to the optimization, and using a sample size of 1 or 2 is sufficient.
On average, the reactive scheme is as efficient as the policy using fixed sample size = 2.
The algorithm automatically detects when it is necessary to use a higher sample size and when, in contrast, a smaller sample size is sufficient.
However, apart from Figure \ref{fig:test10}, the other cases show a scenario in which the optimizer using sample size = 1 is more efficient than the reactive algorithm. 
This happens because the function does not contain a sufficient amount of noise to make estimators too different from expectations, and the method presented in this work requires $n \geq 2$ to provide a statistical guarantee for each comparison.
Furthermore, these experiments show that SSM tends to be too conservative with respect to the reactive scheme, especially as the amount of noise in the objective function increases.
On average, SSM requires higher sample sizes to compare a pair of solutions and take a decision.
Results presented in Figure \ref{fig:test11}, Figure \ref{fig:test12}, Figure \ref{fig:test13} and Figure \ref{fig:test14}, about experiments done on Rastrigin and Rosenbrock functions, are consistent with previous results.
As in all the tests which have been conducted, these plots confirm that SSM is less efficient than the reactive algorithm and its performance worsens considerably as the noise in the objective function increases.


In the second set of experiments, the total revenue computed by HotelSimu's simulator is maximized over a period of 3 months for the following booking scenario.
A small hotel is defined with capacity = 20, booking horizon = 30, maximum number of rooms for a booking = 5 and maximum length of stay for a booking = 14. 
Other details regarding the input data fed to the simulator are not provided, but they correspond to the case of the small hotel defined in \cite{hotelsimu}. However, differently from that example, in this case booking and cancellation curves change every day and acceptance probability is also different for each customer.
Average arrivals, average cancellations and acceptance probabilities are modified so that they can change by $\pm 25\%$ with respect to the original values.
Each experiment is based on 200 macroreplications run with different seeds, and each optimization process uses a number of function evaluations (budget) = 2000.
Lines in the figures represent the mean total revenue of the best configuration at a certain point during the optimization, computed using sample size = 100 to approximate the objective function.
These experiments extend the previous analysis by using RLS with static step size $\in \lbrace 0.2, 0.4, 1 \rbrace$
Also, empirical observation showed that the output of the simulator is normally distributed, and the standard deviation of the output oscillates around 5\% of the mean.

\begin{figure}
\centering
\begin{minipage}{.5\textwidth}
  \centering
  \includegraphics[height=170pt,width=190pt]{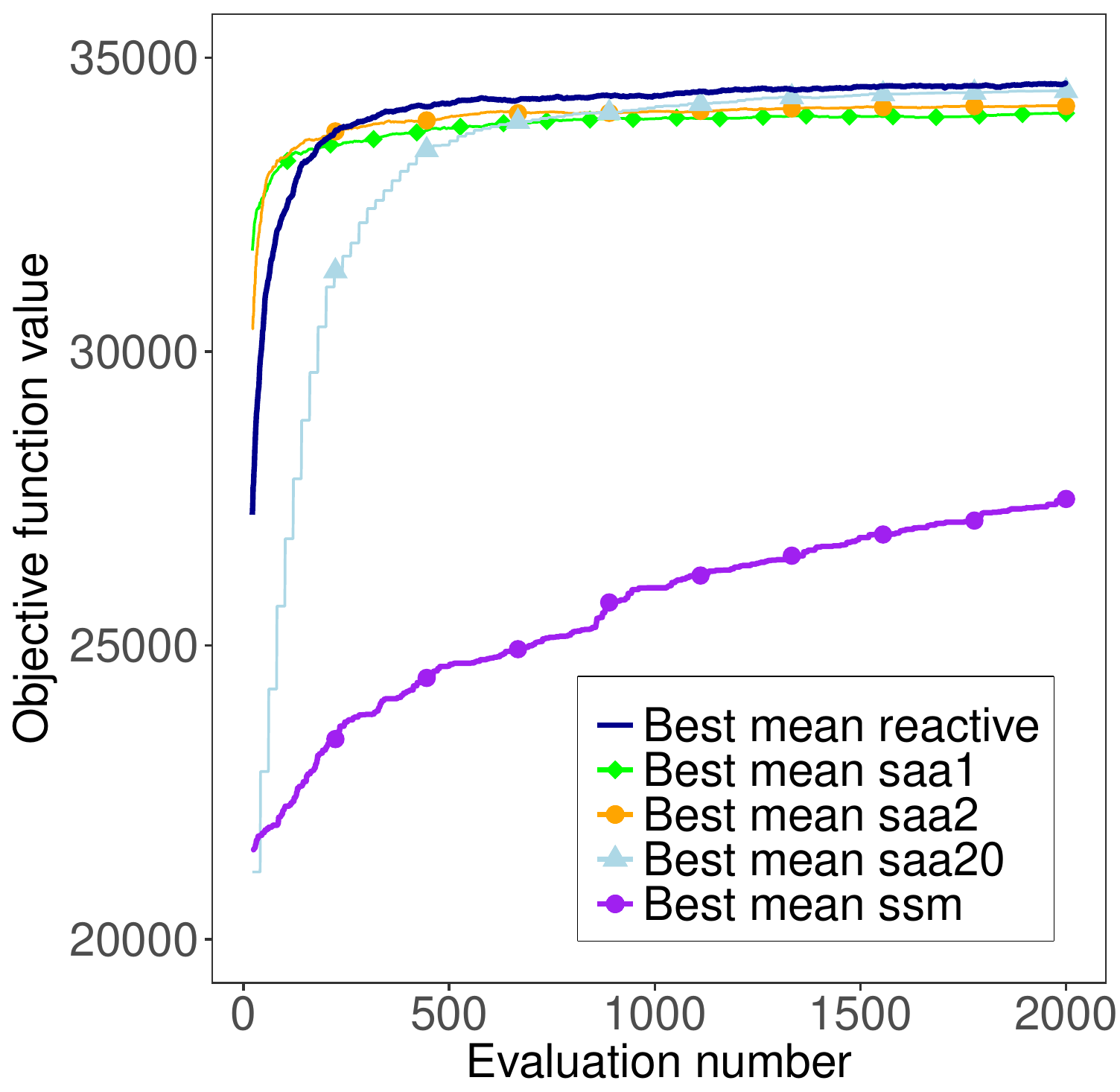}
  \caption{HotelSimu mean trend, step = 0.2.}
  \label{fig:test1}
\end{minipage}%
\begin{minipage}{.5\textwidth}
  \centering
  \includegraphics[height=168pt,width=180pt]{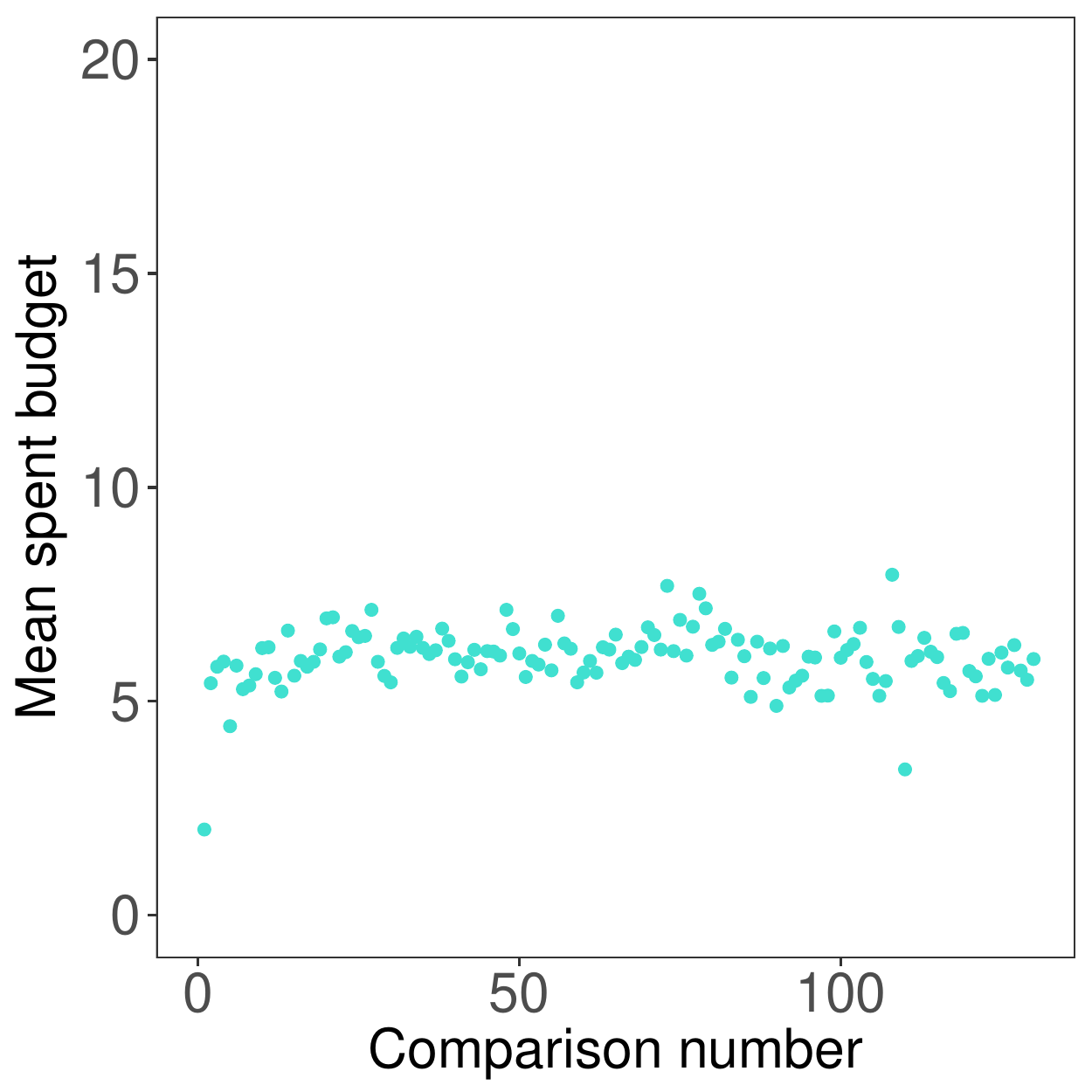}
  \caption{Mean spent budget (reactive), step = 0.2.}
  \label{fig:test2}
\end{minipage}
\end{figure}

\begin{figure}
\centering
\begin{minipage}{.5\textwidth}
  \centering
  \includegraphics[height=170pt,width=190pt]{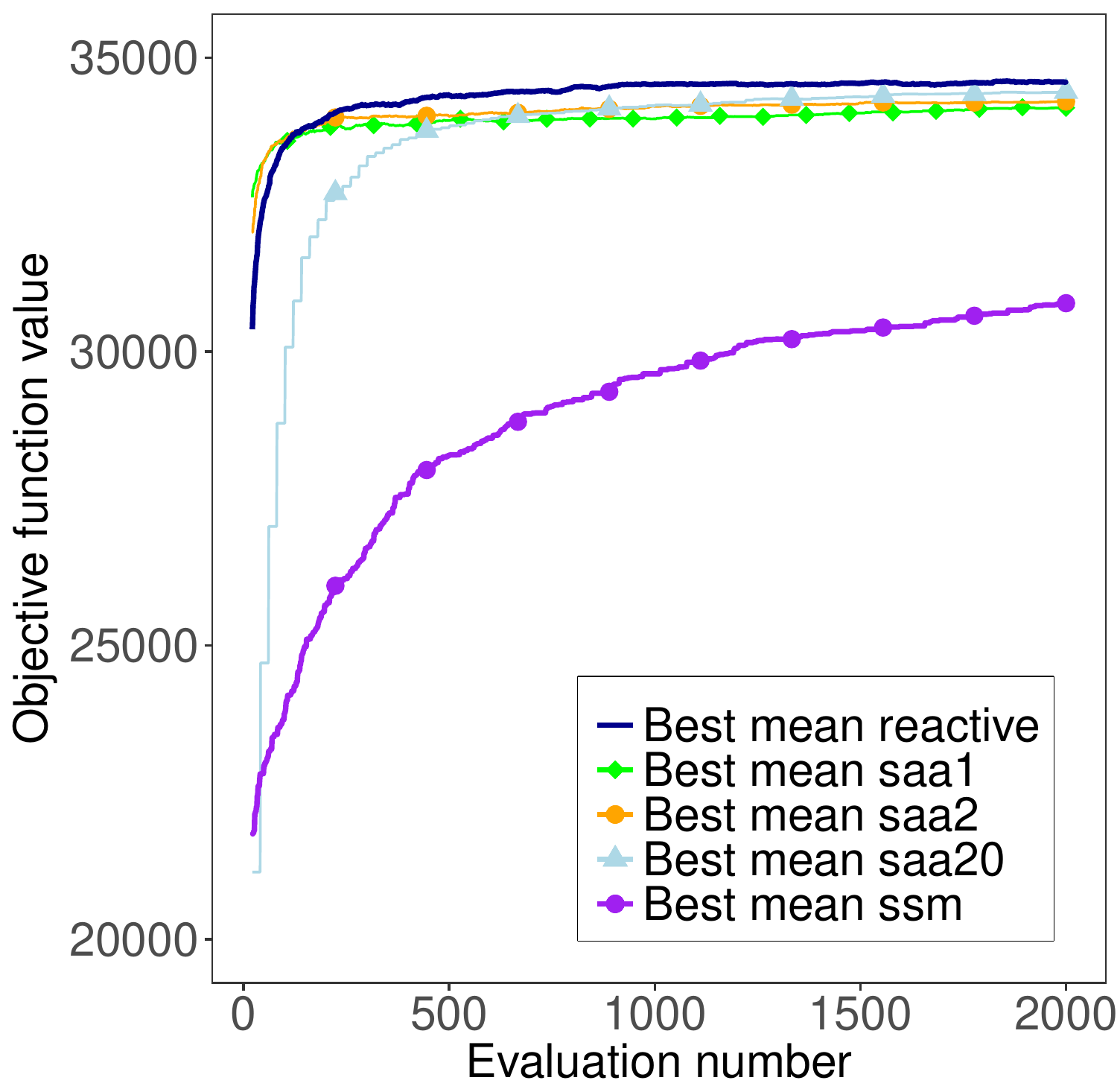}
  \caption{HotelSimu mean trend, step = 0.4.}
  \label{fig:test3}
\end{minipage}%
\begin{minipage}{.5\textwidth}
  \centering
  \includegraphics[width=180pt]{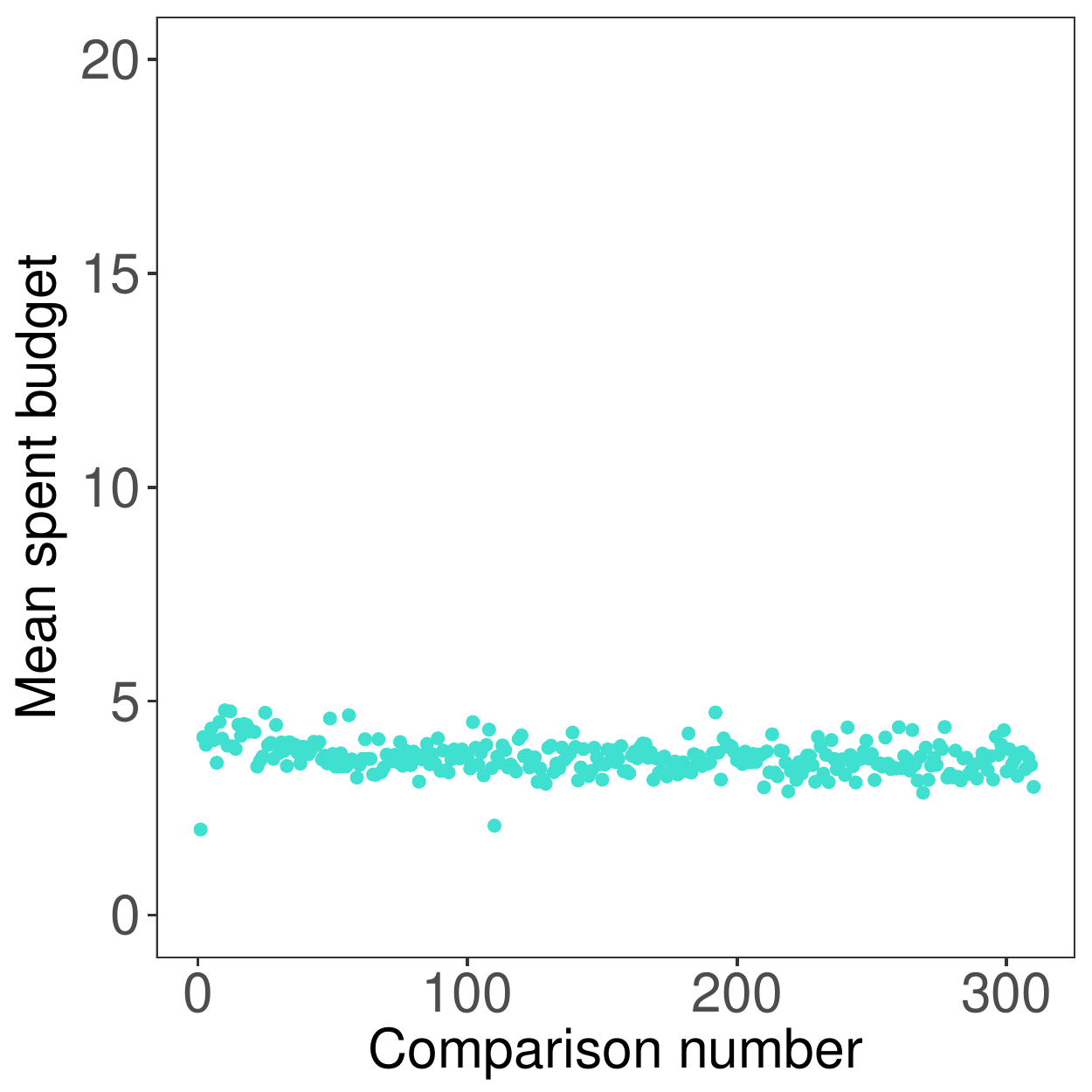}
  \caption{Mean spent budget (reactive), step = 0.4.}
  \label{fig:test4}
\end{minipage}
\end{figure}

\begin{figure}
\centering
\begin{minipage}{.5\textwidth}
  \centering
  \includegraphics[height=170pt,width=190pt]{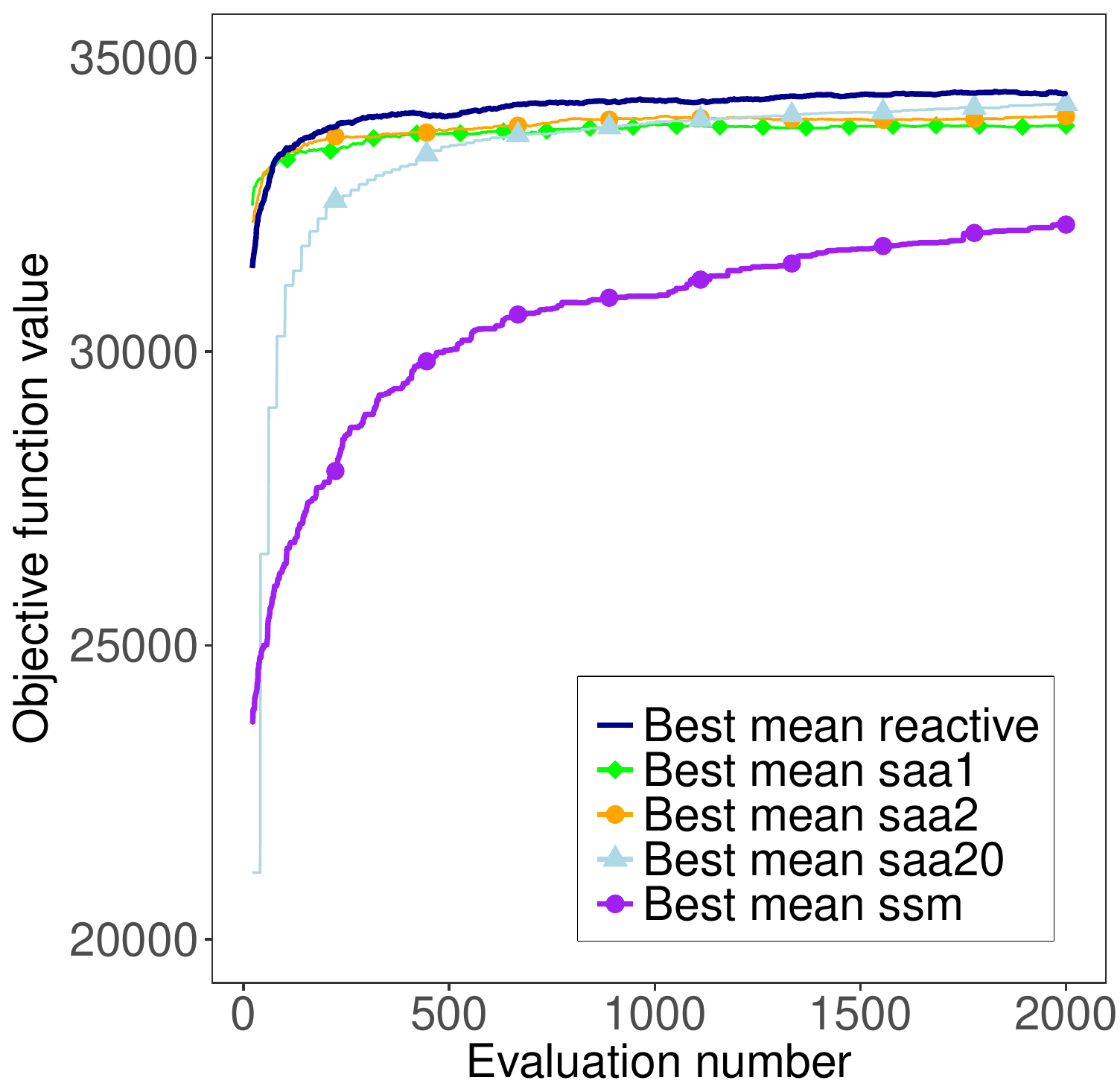}
  \caption{HotelSimu mean trend, step size = 1.}
  \label{fig:test5}
\end{minipage}%
\begin{minipage}{.5\textwidth}
  \centering
  \includegraphics[width=180pt]{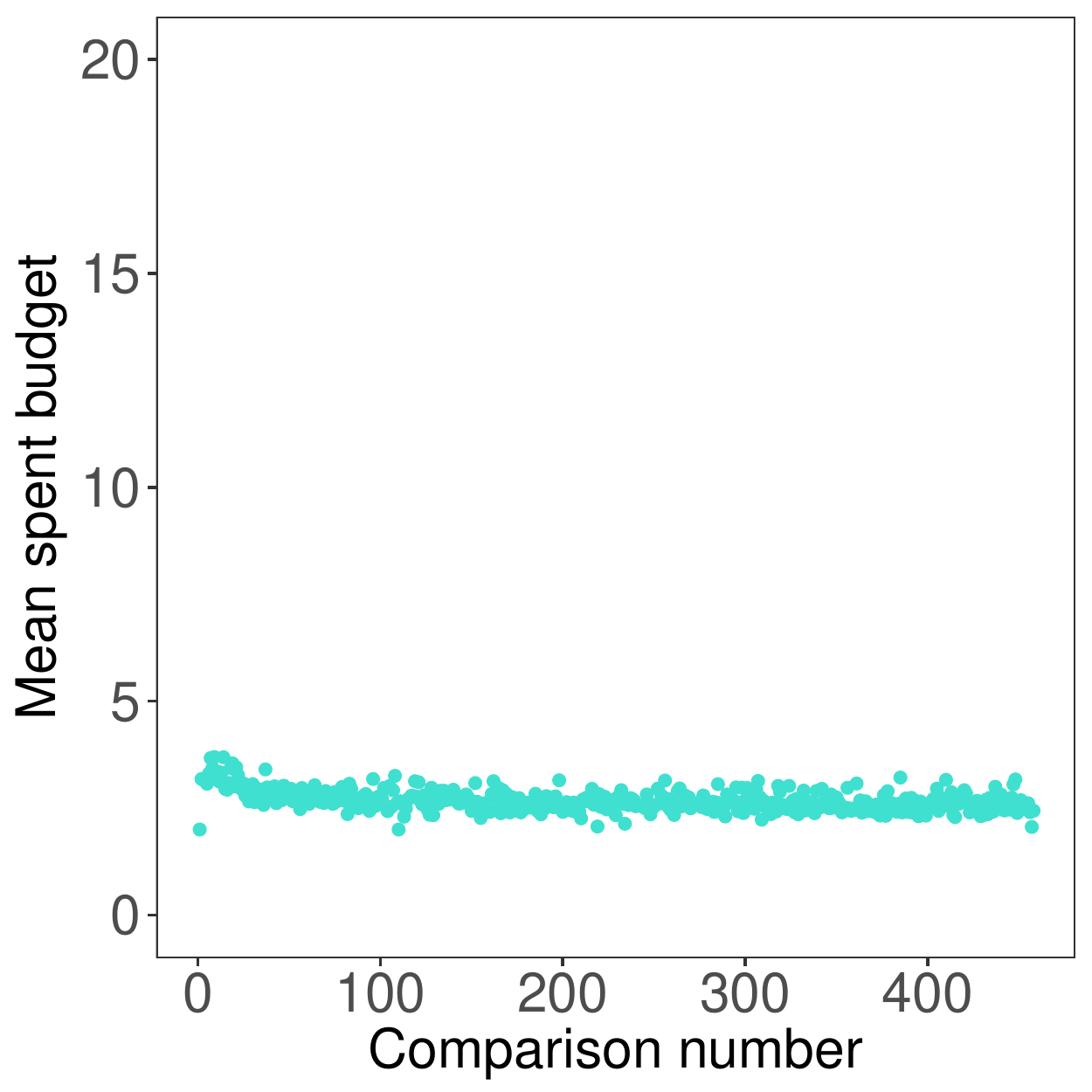}
  \caption{Mean spent budget (reactive), step = 1.}
  \label{fig:test6}
\end{minipage}
\end{figure}

Figure \ref{fig:test1}, Figure \ref{fig:test3} and Figure \ref{fig:test5} show the convergence of RLS optimizers with different step sizes. 
Results are consistent with previous observations, but they also show that the locality of the search policy influences considerably the performance of SSM. 
As the neighborhood of the local search policy becomes smaller, and so solutions visited during the search are more similar, the efficiency of SSM decreases. 
When using SSM, RLS policies obtain less efficient results with respect to pure random search.
The reactive algorithm is also subject to these problems, but the impact that the locality of the search has on it is not as large as the effect obtained on SSM.


Figure \ref{fig:test2}, Figure \ref{fig:test4} and Figure \ref{fig:test6}  show the mean sample size used for comparisons throughout the optimizations when using the reactive algorithm. As the step size increases, and the optimization process becomes more global, the mean sample size used for comparisons decreases. 
This happens because on average comparisons involve configurations which are more different, because they can be drawn from distant areas of $\Theta$. 
This also has an impact on the exploration of the heuristic algorithm, because the more evaluations are needed on average for comparing configurations, the less budget is available for proposing new candidates.

\section{Conclusions}
This work introduces a novel reactive sample size algorithm based on parametric tests and IZ selection, which can be used for improving the efficiency and robustness of heuristic optimization techniques used in the context of simulation-based optimization.
The algorithm is evaluated by using a real simulation-based optimization tool for hotel revenue management and three deterministic functions, extended with multiple levels of normally distributed noise.
The performance of the reactive algorithm is compared with a more naive scheme which constantly uses a fixed sample size, and with SSM.

Experimental results show that the presented algorithm improves the convergence of heuristic optimization algorithms based on RLS, and it is more efficient than SSM in all experiments. 
Also, by using statistically significant comparisons during the heuristic search, it is not necessary to run preliminary experiments in order to find \textit{a priori} the minimum sample size to use throughout the optimization.
By modifying the parameters of the algorithm, comparisons with different levels of statistical significance can be obtained.
In fact, because of the simulation scenario, in some cases it might be very important to avoid taking wrong decisions; by increasing $\alpha_{req}$ and $\beta_{req}$, more robust results can be obtained. 
Also, $\delta$ can be increased in order to obtain less robust results by taking heuristic decisions more often.
The parameters of the algorithm give to the user the possibility to find a tradeoff between efficiency and robustness, according to the needs defined by the problem which is simulated.

\bibliographystyle{unsrt}  
\bibliography{references}

\end{document}